# Optimizing Simulations with Noise-Tolerant Structured Exploration

Krzysztof Choromanski, Atil Iscen, Vikas Sindhwani, Jie Tan and Erwin Coumans

*Abstract*— We propose a simple drop-in noise-tolerant replacement for the standard finite difference procedure used ubiquitously in blackbox optimization. In our approach, parameter perturbation directions are defined by a family of structured orthogonal matrices. We show that at the small cost of computing a Fast Walsh-Hadamard/Fourier Transform (FWHT/FFT), such structured finite differences consistently give higher quality approximation of gradients and Jacobians in comparison to vanilla approaches that use coordinate directions or random Gaussian perturbations. We find that trajectory optimizers like Iterative LQR and Differential Dynamic Programming require fewer iterations to solve several classic continuous control tasks when our methods are used to linearize noisy, blackbox dynamics instead of standard finite differences. By embedding structured exploration in a quasi-Newton optimizer (LBFGS), we are able to learn agile walking and turning policies for quadruped locomotion, that successfully transfer from simulation to actual hardware. We theoretically justify our methods via bounds on the quality of gradient reconstruction and provide a basis for applying them also to nonsmooth problems.

## I. INTRODUCTION

Recent years have witnessed rapid increase in sophistication and realism in physics engines [9], [1] and 3D renderers, a trend complemented by an ever-growing capacity to run massive-scale simulations [20] in distributed environments. These developments are raising expectations that robot motor skills and control policies learnt at scale in complex virtual worlds, possibly tweaked with small amounts of real-world experience, will successfully transfer to reality. In this context, we revisit arguably the oldest and simplest methodology for derivative-free optimization [8] and direct policy search [19], [10]: the ubiquitous finite difference method.

Despite their simplicity, finite difference methods remain the method of choice in the simulation optimization [4] literature, particularly when simulations are computationally inexpensive to run, as is increasingly the case. Recent work [20] shows that randomized variants of the finite difference method are surprisingly competitive with state of the art Deep Reinforcement Learning (RL) algorithms on the most challenging benchmark environments in Mujoco [1]. The notorious data inefficiency of finite difference schemes is mitigated by the ease with which they can be parallelized, and their simplicity is complemented by their flexibility in terms of being immediately pluggable into mature numerical solvers; in addition, they require no special treatment to

Krzysztof Choromanski, Atil Iscen and Vikas Sindhwani are in Google Brain team in New York, USA. Jie Tan and Erwin Coumans are in Google Brain team in Mountain View, CA, USA. {kchoro, atil, sindhwani, jietan, erwincoumans}@google.com

handle sparse rewards and long time horizons when used for policy optimization. However, despite many success stories [15], [18], [11], their brittleness in the face of noise and, until recently [17], lack of theoretical basis for optimizing non-smooth objectives has remained a concern.

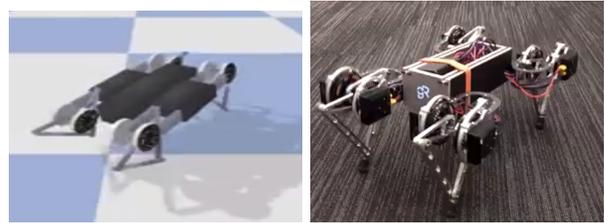

Fig. 1. Minitaur (from Ghost Robotics) executing a turning maneuver: in Bullet simulator (left), in reality (right), using our methods.

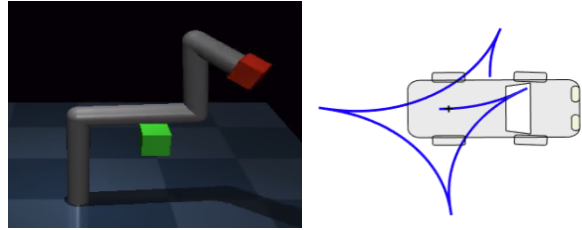

Fig. 2. 5DOF Arm Control in Mujoco simulator using TROSS [21] (left), Car Parking with Iterative LQR [16], [22] (right). We show improvements in trajectory optimization in these tasks using proposed methods.

*Preliminaries:* In this paper, we are interested in solving general simulation optimization problems of the form,

$$\mathbf{x}^* = \arg\min_{\mathbf{x} \in \Omega \subseteq \mathbb{R}^n} f(\mathbf{x})$$

Here, $\mathbf{x} \in \mathbb{R}^n$ are simulation optimization variables, possibly constrained to live in a subset $\Omega$ (e.g., bound constraints). The objective function, $f$, is a scalar-valued function of the internal states encountered during the simulation. In RL policy optimization, $f$ is negative of expected return over the duration of the simulation where the robot interacts with the environment using an open-loop or closed-loop control policy parameterized by $\mathbf{x}$. As in model-free RL settings, we assume that the derivatives of $f$ are unavailable since the underlying physics engine is treated as a blackbox. Furthermore, every time $f$ is evaluated at $\mathbf{x}$, we actually get back $f(\mathbf{x}) + \varepsilon(\mathbf{x})$ where $\varepsilon(\mathbf{x})$ is noise.

To set the stage, we recall the basics of the finite difference method for approximating gradients of such a blackbox assuming $f$ is differentiable. By Taylor approximation, the gradient characterizes the rate of change in $f$ at a step $\delta$ ($0 < \delta \ll 1$) along any direction $\mathbf{d} \in \mathbb{R}^n$, as follows,

$$\nabla f(\mathbf{x}_0)^T \mathbf{d} \approx \frac{1}{\delta}\left[f(\mathbf{x}_0 + \delta \mathbf{d}) - f(\mathbf{x}_0)\right]. \quad (1)$$

Hence, given a set of $q$ directions $\mathbf{d}_1 \ldots \mathbf{d}_q$ in $\mathbb{R}^n$, and associated finite difference measurements $m_i = \frac{1}{\delta}\left[f(\mathbf{x}_0 + \delta \mathbf{d}_i) - f(\mathbf{x}_0)\right] \in \mathbb{R}$, one can recover the gradient by solving a least squares problem,

$$\widehat{\nabla f}(\mathbf{x}_0) = \arg\min_{\mathbf{z} \in \mathbb{R}^n} \sum_{i=1}^{q} (\mathbf{z}^\top \mathbf{d}_i - m_i)^2 = \|\mathbf{M}\mathbf{z} - \mathbf{m}\|_2^2$$

where $\mathbf{M} = [\mathbf{d}_1 \ldots \mathbf{d}_q]^\top$ and $\mathbf{m} = [m_1 \ldots m_q]^\top$. We will interchangeably refer to $\mathbf{d}_i$ as exploration or perturbation directions. In this paper, we study the most common and classical, $q = n$, case though in general both the underdetermined ($q < n$) and overdetermined ($q > n$) cases can also be well-motivated. When $q = n$, assuming $\mathbf{M}$ is non-singular, we need to solve the linear system, $\mathbf{M}\mathbf{z} = \mathbf{m}$, to obtain an estimate $\mathbf{z}$ of the gradient. In the derivative-free optimization literature this estimate is also referred to as the *simplex gradient* associated with $\mathbf{M}$ [8]. When perturbation directions $\mathbf{d}_i$ are elements $\mathbf{e}_i$ of the canonical basis (the vector of all zeros except for a 1 in the $i^{th}$ entry), $\mathbf{M}$ is the identity matrix and the estimator is immediate,

$$\widehat{\nabla f}(\mathbf{x}_0)_i = \frac{1}{\delta}\left[f(\mathbf{x}_0 + \delta \mathbf{e}_i) - f(\mathbf{x}_0)\right] \quad (2)$$

This case corresponds to the default and most popular version of the finite difference method [23].

The randomized finite difference gradient estimator studied in [20], [17] has the following form,

$$\widehat{\nabla f}(\mathbf{x}_0) = \sum_{i=1}^{n} \frac{1}{\delta}\left[f(\mathbf{x}_0 + \delta \mathbf{g}_i) - f(\mathbf{x}_0)\right] \mathbf{g}_i \quad (3)$$

where the perturbation directions $\mathbf{g}_i$ are drawn from a standard multivariate Gaussian distribution. This estimator may be justified through the notion of *Gaussian smoothing* [17]. For a function $f : \mathbb{R}^n \to \mathbb{R}$ and $\delta > 0$ we define its *Gaussian smoothing* as:

$$f_\delta(\mathbf{x}) = \mathbb{E}_{\mathbf{g} \sim \mathcal{N}(0, \mathbf{I})}[f(\mathbf{x} + \delta \mathbf{g})].$$

The gradient of the Gaussian smoothing of $f$ has the following form [17],

$$\nabla f_\delta(\mathbf{x}) = \mathbb{E}\left[\frac{f(\mathbf{x} + \delta \mathbf{g}) - f(\mathbf{x})}{\delta} \mathbf{g}\right]. \quad (4)$$

Hence, the randomized estimator in Eqn. 3 may be seen as a Monte-Carlo approximation to the gradient of the Gaussian smoothing of $f$. This also provides a justification of using the estimator even for non-smooth problems (in fact, see [17] for convergence rates of finite-difference based gradient descent in the non-convex non-smooth setting).

*Our Contributions*: We propose a simple drop-in replacement for the standard finite difference procedure where the perturbation directions are rows of certain deterministic or randomized structured orthogonal matrices. All our estimators turn out to have the form,

$$\widehat{\nabla f}(\mathbf{x}_0) = \sum_{i=1}^{n} \frac{1}{\delta}\left[f(\mathbf{x}_0 + \delta \mathbf{h}_i) - f(\mathbf{x}_0)\right] \mathbf{h}_i \quad (5)$$

A canonical example is when perturbation directions $\mathbf{h}_i$ are rows of a Hadamard matrix. Since Hadamard matrices admit fast matrix-vector products using FWHT, the structured gradient estimator in Eqn. 5 can be computed in $O(n \log n)$ time. In section II, we describe a broader family of structured matrices, termed *balanced spinners*, that can be used to define perturbation directions. We provide a theoretical justification for such estimators in the presence of noise. In section III, we consider three classic continuous control problems: Cartpole and Acrobot balancing, and a car maneuvering problem studied in [22] (see Figure 2, right). We inject increasing levels of noise in the underlying dynamics and study the behavior of Iterative LQR [16] using different finite difference approximations of the state and control Jacobians of the dynamics. We find that the proposed methods consistently outperform standard finite differences and the random Gaussian version [20], [17] of Eqn. 3. We also obtain more precise target reaching motions with a 5-dof arm simulated in Mujoco [1] (see Figure 2, left), using our methods in conjunction with the TROSS [21] trajectory optimizer. In section IV, we apply our methods to quadruped locomotion tasks. In particular, we learn open loop policies for forward walking and turning from Bullet [9] simulations of the Minitaur [2] quadruped robot. For optimization, we use L-BFGS-B [6], [24] a limited-memory quasi-Newton method for bound-constrained optimization, where gradients are computed using our structured finite difference method. This approach in particular may be seen as structured extension of Implicit Filtering methods [8], [14]. Learnt policies succesfully transfer to the real Minitaur (Fig. 1).

## II. FILTERING OUT NOISE WITH BALANCED SPINNERS

In this section we formally introduce the family of structured matrices that we call *balanced spinners* whose rows define perturbation directions for constructing finite difference approximations. We describe both deterministic as well as randomized constructions; the latter offer stronger guarantees regarding the quality of the reconstructed gradients. As we show in the experimental section, the randomized variants can also be chained together to create the so-called *multispinners* that in practice often work better. We also explain why our approach can be used also when $f$ is not differentiable: in that case the gradient of the associated Gaussian smoothing [17] is estimated. For proofs, please see the Appendix.

### A. Noise resilient $(\alpha, \beta)$-balanced spinners

We are interested in the following class of matrices:

***Definition 2.1 (($\alpha,\beta$)-balanced spinners):*** We say that a matrix $\mathbf{M} = \{M_{i,j}\}_{i,j=1,...,n} \in \mathbb{R}^n$ is an $(\alpha,\beta)$-balanced spinner for $0 \leq \alpha, \beta \leq 1$ if it is invertible and the following is true:
- $\|\mathbf{m}_i\|_2 \geq \alpha\sqrt{n}$ and $\sup_{i,j \in \{1,...,n\}} |M_{i,j}| \leq 1$,
- $|\mathbf{m}_i^\top \mathbf{m}_j| \leq (1-\beta) \min_{i=1,...,n} \|\mathbf{m}_i\|_2$ for $1 \leq i < j \leq n$,

where $\mathbf{m}_i^\top$ stands for the $i$th row of $\mathbf{M}$.

It will turn out that better quality balanced spinners are characterized by $\beta \approx 1$. In other words, we want the rows of $\mathbf{M}$ to be nearly orthogonal.

To apply matrices from this class, we will choose as $n$ perturbation directions $\mathbf{d}_1, ..., \mathbf{d}_n$, the rows of a fixed balanced spinner. Assume that each finite difference measurement is corrupted by noise $\eta_i$, denote by $\eta$ the noise vector $\eta = (\eta_1, ..., \eta_n)^\top$.

We denote by $\mathbf{m} = (m_1, ..., m_n)^\top$ the vector of finite difference measurements and assume that the perturbation directions are taken from the unit hypercube in the $L_\infty$-norm, i.e. to perform the finite difference with a fixed step size $\delta$, we do not perturb each dimension by more than 1. We give the following guarantee.

***Theorem 2.1:*** Assume that the perturbation directions $\mathbf{d}$ are chosen from the $L_\infty$ cube: $\|\mathbf{d}\|_\infty \leq 1$ and the associated finite difference measurements are bounded in the $L_2$-norm by $\Delta$. Let $\mathbf{z}$ be the unique solution to the following linear system: $\mathbf{Mz} = \mathbf{m}$, where $\mathbf{m}$ is measurements vector and the finite difference directions are rows of the $(\alpha,\beta)$-balanced spinner $\mathbf{M}$. Then the following is true:

$$\|\mathbf{z} - \nabla f(\mathbf{x}_0)\|_2 \leq \left(\frac{(1-\beta)\sqrt{n}}{\alpha^2 \sigma_{\min}(\mathbf{M})} + \frac{\|\mathbf{M}^\top\|_2}{\alpha^2 n}\right)\Delta,$$

where $\sigma_{\min}(\mathbf{M})$ stands for the minimal singular value of $\mathbf{M}$.

We do not require exact orthogonality to provide gradient approximation guarantees (as long as $\beta \approx 1$). However, of particular interest in this paper are $(1,1)$-balanced spinners for which we have the following tighter reconstruction guarantee as a corollary of the theorem above.

***Corollary 2.1:*** Under the assumptions of Theorem 2.1, if $\mathbf{M}$ is a $(1,1)$-balanced spinner, we have,

$$\|\mathbf{z} - \nabla f(\mathbf{x}_0)\|_2 \leq \frac{1}{\sqrt{n}}\Delta,$$

In comparison, standard finite difference corresponding to identity matrices incurs larger error.

***Remark 2.1 (structured vs standard finite difference):*** The bound obtained by applying standard finite difference algorithm leads to the larger $L_2$-error of the gradient reconstruction (it corresponds to taking: $\alpha = \frac{1}{\sqrt{n}}$ and $\beta = 1$).

Several structured matrices admitting fast matrix-vector products as well as compact storage can be shown to be $(1,1)$-balanced spinners. We next list some explicit constructions that are also used in our experiments.

**Hadamard matrices:** Every matrix of the form $\mathbf{M} = \mathbf{B}_1 \otimes ... \otimes \mathbf{B}_l$, where $\mathbf{B}_1 = ... = \mathbf{B}_l \in \{-1, +1\}^{2 \times 2}$, $\mathbf{B}_1[i][j] = 1$ if $i \neq 1$ or $j \neq 1$, $\mathbf{B}_1[1][1] = -1$ and $\otimes$ stands for the Kronecker product, is an $(1,1)$-balanced spinner from $\mathbb{R}^{2^l \times 2^l}$. We call $\mathbf{M}$ a *Hadamard matrix*.

**Quadratic residue matrices:** Another family of $(1,1)$-balanced spinners is obtained by a the following construction from graph theory. Denote by $p$ prime number of the form $p = 4t + 3$ for $t \in \mathbb{N}$. A *tournament $T$* is a complete directed graph without multi-edges. *Quadratic residue tournament $T_p$* for $p$ as above is a tournament on $p$ vertices $\{0, 1, ..., p-1\}$, where there exists an edge from $i$ to $j$ for $0 \leq i, j \leq p-1$, iff $i - j$ is a *quadratic residue modulo $p$*, i.e if there exists $a \in \mathbb{N}$ such that $i - j \equiv a^2 \pmod{p}$.

Consider an adjacency matrix $\mathbf{Q}_p \in \{-1, 1\}^{p \times p}$, of the quadratic residue tournament $T_p$ i.e. a matrix such that for $i \neq j$, $\mathbf{Q}_p[i][j] = 1$ if there exists an edge from $i$ to $j$ in $T_p$, $\mathbf{Q}_p[i][j] = -1$ otherwise and $\mathbf{Q}_p[i][i] = 1$. Denote by $\mathbf{Q}_p^* \in \mathbb{R}^{(p+1) \times (p+1)}$ a matrix defined as: $\mathbf{Q}_p^*[i][j] = -1$ if $i = 0$ or $j = 0$ and $\mathbf{Q}_p^*[i][j] = \mathbf{Q}_p^\top[i-1][j-1]$ otherwise. We call $(\mathbf{Q}_p^*)^\top$ a *quadratic residue matrix*. The following is true.

***Lemma 2.1:*** Every matrix $(\mathbf{Q}_p^*)^\top$ is an $(1,1)$-balanced spinner.

The above lemma is a direct consequence of the definition of $\mathbf{Q}_p$ and the following well-known result showing that columns of $\mathbf{Q}_p$ are nearly-orthogonal (see: [3]):

***Lemma 2.2:*** Every two different columns $\mathbf{c}_i$ and $\mathbf{c}_j$ of $\mathbf{Q}_p$ satisfy: $\mathbf{c}_i^\top \mathbf{c}_j = -1$.

**Randomized constructions:** The above two constructions are discrete, but we can easily randomize them. Note first the following remark.

***Remark 2.2:*** For any fixed $0 \leq \alpha, \beta \leq 1$, the family $\mathcal{M}_{\alpha,\beta} \subseteq \mathbb{R}^n$ of $(\alpha,\beta)$-balanced spinners is closed under multiplying from the right by the diagonal matrix $\mathbf{D} \in \mathbb{R}^n$ with nonzero entries taken from the 2-element set $\{-1, +1\}$.

Now, by Remark 2.2, we can conclude that if $\mathbf{W} \in \mathbb{R}^n$ is of the form $\mathbf{W} = \mathbf{MD}$, where $\mathbf{D}$ is as above and $\mathbf{M}$ is a $(1,1)$-balanced spinner, then $\mathbf{W}$ is an $(1,1)$-balanced spinner. In particular, matrices of the form $\mathbf{HD}$ (called later Hadamard random) or $(\mathbf{Q}_p^*)^\top \mathbf{D}$, where $\mathbf{H}$ and $\mathbf{Q}_p^*$ are as above, are also $(1,1)$-balanced spinners

It turns out that for these randomized constructions, we can obtain much stronger guarantees and measure the reconstruction error not only in $L_2$, but also in the $L_\infty$-norm. We show the result below for Hadamard matrices, but it can be generalized for other $(\alpha,\beta)$-balanced spinners. The following is true:

*Theorem 2.2:* Replace matrix **M** in Theorem 2.1 by a matrix **HD**, where **D** is a random diagonal matrix with entries taken independently at random from $\{-1,+1\}$ and **H** is Hadamard. Let $g(n) : \mathbb{N} \to \mathbb{R}$ be any increasing function and assume that $\|\eta\|_\infty \leq \sqrt{\frac{n}{\log(n)}} \frac{1}{g(n)}$. Then the following is true with probability at least $1 - 2ne^{-\frac{g(n)\log(n)}{2}}$:

$$\|\mathbf{z} - \nabla f(\mathbf{x}_0)\|_\infty \leq \frac{1}{\sqrt{g(n)}},$$

where **z** stands for the approximate gradient.

*Fast Computations:* Gradient reconstruction with perturbation directions defined by **H**, **HD** or $(\mathbf{Q}_p^*)^\top$ can be done fast because of the following observations: due to orthogonality, for these matrices we have $\mathbf{M}^{-1} = \frac{1}{n}\mathbf{M}^\top$; being structured, these matrices admit fast matrix-vector multiplication via FWHT or FFT [12]. Note that by using FWHT or FFT, one does not even need to store matrix **M** explicitly, so gradient reconstruction can be done in $O(n)$ space (storage of the measurements vector).

Note that we do not need any assumptions regarding independence of different elements of the noise vector $\eta$. In particular, our result covers the adversarial noise model, where different elements of $\eta$ may depend in a convoluted way on other elements.

### B. Approximate gradients of Gaussian Smoothings

We now justify the use of our methods even for nonsmooth problems. Suppose the blackbox $f$ is not differentiable. Consider its Gaussian smoothing,

$$f_\delta(\mathbf{x}) = \mathbb{E}_{\mathbf{g} \sim \mathcal{N}(0,\mathbf{I})}[f(\mathbf{x} + \delta \mathbf{g})].$$

The gradient of the Gaussian smoothing of $f$ has the following form [17],

$$\nabla f_\delta(\mathbf{x}) = \mathbb{E}[\frac{f(\mathbf{x}+\mu \mathbf{g}) - f(\mathbf{x})}{\delta}\mathbf{g}]. \quad (6)$$

Note that gradient approximation $\widehat{\nabla f}(\mathbf{x}_0)$ obtained by the finite difference method with perturbation directions drawn from a matrix **M** is of the form:

$$\nabla f(\mathbf{x}_0)_{\text{fd}} = \mathbf{M}^{-1}\mathbf{m}, \quad (7)$$

where $\mathbf{m} \in \mathbb{R}^n$ is the measurements vector. Notice that for $\mathbf{M} = \mathbf{H}$, $\mathbf{M} = \mathbf{HD}$, $\mathbf{M} = (\mathbf{Q}_p^*)^\top$, $\mathbf{M} = (\mathbf{Q}_p^*)^\top \mathbf{D}$ and normalized version **M** of $\mathbf{W} = \mathbf{HD}_1 \cdot ... \cdot \mathbf{HD}_k$ and $\mathbf{W} = (\mathbf{Q}_p^*)^\top \mathbf{D}_1 \cdot ... \cdot (\mathbf{Q}_p^*)^\top \mathbf{D}_k$, matrix $\frac{1}{\sqrt{n}}\mathbf{M}$ is an isometry.

Thus we obtain: $\mathbf{M}^{-1} = \frac{1}{n}\mathbf{M}^\top$. Note also that the $i^{th}$ entry $m_i$ of **m** is of the form: $m_i = \frac{f(\mathbf{x}_0 + \delta \mathbf{d}_i) - f(\mathbf{x}_0)}{\delta}$, where $\mathbf{d}_i$ stands for the $i^{th}$ row of **M**. Therefore we get:

$$\widehat{\nabla f}(\mathbf{x}_0) = \frac{1}{n}\sum_{i=1}^n \frac{f(\mathbf{x}_0 + \delta \mathbf{d}_i) - f(\mathbf{x}_0)}{\delta}\mathbf{d}_i, \quad (8)$$

where the right hand side is exactly a Monte-Carlo approximation of $\nabla f_\mu(\mathbf{x}_0)$ with Gaussian vectors $\mathbf{g}_1, ..., \mathbf{g}_n$ replaced by $\mathbf{d}_1, ..., \mathbf{d}_n$. The effectiveness of the finite difference method with the above matrices follows now from the fact that the distribution of rows $\mathbf{d}_i$ resembles Gaussian distribution. Using results from [5], it can be shown (see [5], proof of Theorem 1) that rows of the structured random matrices considered here are close to the Gaussian vector $\mathbf{g} \in \mathbb{R}^n$ in a certain technical sense whose details we omit in this paper.

We conjecture that the fact that the above structured estimators are particularly good follows from their low variance and might be related to the recent discovery that certain kernel estimators based on structured **orthogonal** matrices strictly outperform accuracy-wise their baseline counterparts built from Gaussian unstructured matrices (see: [7]).

## III. DERIVATIVE-FREE TRAJECTORY OPTIMIZATION

We benchmark various finite difference approximations in the context of solving trajectory optimization problems of the form,

$$\arg\min_{\mathbf{u}_0...\mathbf{u}_{T-1}} \sum_{t=0}^T c_t(\mathbf{x}_t, \mathbf{u}_t) \ + \ c_T(\mathbf{x}_T)$$
$$\mathbf{x}_{t+1} = f(\mathbf{x}_t, \mathbf{u}_t)$$

where $\mathbf{x}_{t+1} = f(\mathbf{x}_t, \mathbf{u}_t)$ denotes a discrete-time dynamical system with states $\mathbf{x}_t \in \mathbb{R}^n$ and control inputs $\mathbf{u}_t \in \mathbb{R}^m$. Above, $T$ is the optimization horizon, $c_t$ and $c_T$ are stagewise and terminal cost functions. We are interested in the setting where we only have noisy, blackbox access to the dynamics $f$. Differential Dynamic Programming (DDP) [13] and Iterative Linear Quadratic Regulator (iLQR) [16], variations of Newton's method, are among the most effective methods for solving such problems. At each iteration, a time-varying Linear Quadratic Regulator (TV-LQR) subproblem is solved, by linearizing the dynamics and constructing a quadratic approximation to the cost function, yielding a direction along which the current trajectory is updated via line search. Finite-difference Jacobian approximations are typically used in the dynamics linearization step. The optimal controls can be expressed in the form of locally linear feedback control policies. Extensions of DDP/iLQR for handling bound constraints on control variables were proposed in [22]. An ADMM based extension called TROSS that can handle nonsmooth costs and constraints was proposed in [21].

In this section, our goal is to embed various finite difference approximations in these solvers and study the resulting optimization behavior, while injecting noise in the dynamics.

### A. Car parking with Control-limited DDP *[22]*

In this task, studied in [22], a car is described by a 4-dimensional state vector: $\mathbf{x} = (p_x, p_y, \theta, v)$ where $(p_x, p_y)$ is the position of a point midway between the back wheels, $\theta$ is the angle of the car relative to the $x$-axis, and $v$ is the velocity of the front wheels. The front wheel angle $\omega$ and the front wheel acceleration $a$ are the control inputs, with limits $\pm 0.5$ radians, and $\pm 2.0$ meters-per-second-square respectively. The car needs to be maneuvered from a starting state $p_x = 1, p_y = 1, \theta = \frac{3\pi}{2}, v = 0$ to a parking goal state, $p_x = 0, p_y = 0, \theta = 0, v = 0$. The ILQR parking trajectory is shown in Figure 2 (right).

*1) Convergence Rates:* In Figure 3 we present a comparison of several finite difference methods in terms of ILQR cost-reduction as a function of iterations. It is clear that proposed methods based on $(1,1)$-balanced spinners are characterized by the fastest convergence rate. In fact, the gains are visible (not shown) even in the noiseless setting. ILQR shows markedly slower convergence with Gaussian perturbations, the methods of [20], [17], particularly in the early stages of optimization.

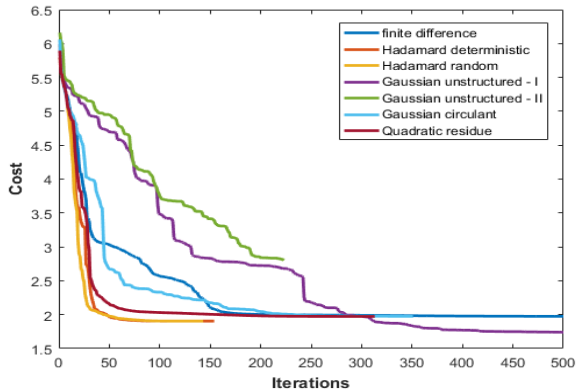

Fig. 3. Comparison of different finite difference methods ("finite difference" stands for the standard finite difference method) for the car parking problem. The ones based on randomized $(1,1)$-balanced spinners are superior to others. Gaussian noise with stddev $= 0.0001$ is added to all finite difference calculations. 'Gaussian unstructured I' and 'Gaussian unstructured II' correspond to two different convergence rate profiles obtained if $\mathbf{M} = \mathbf{G}$, where $\mathbf{G}$ is a Gaussian matrix.

*2) Time complexity:* In the left subfigure of Figure 4 we show that even though time complexity per one gradient computation is larger for the structured finite difference method than the canonical one ($O(n\log(n))$ versus $O(n)$), since the obtained gradients are much more accurate, the total time till convergence is up to two times shorter than for the canonical version. Structured approach is also faster than the one based on unstructured Gaussian matrices.

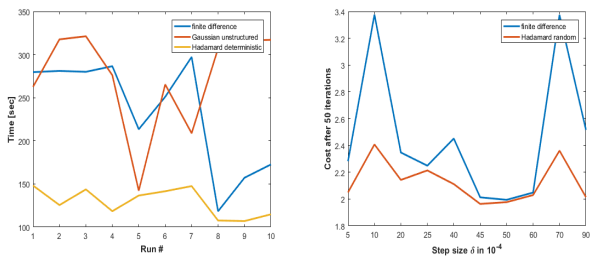

Fig. 4. On the left: comparison of the total running time of the car parking algorithm for structured, unstructured Gaussian and standard finite difference for 10 runs corresponding to independent choices of the random Gaussian noise added, each of stddev $= 0.0001$. On the right: comparison of the cost obtained after 50 iterations for different step sizes $\delta$ for structured randomized finite difference and the standard one.

Figure 4 presents results for $\mathbf{H}$ and $\mathbf{HD}$, but similar results were obtained for quadratic residue matrices.

*3) Robustness to different step size parameters $\delta$:* In the right subfigure of Figure 4 we show how the accuracy of the algorithm changes with different values of the step size parameter $\delta$. We see that the structured approach consistently leads to lower costs than standard finite difference and furthermore, is characterized by much lower variance.

*4) Acrobot and Cartpole Balacing with iLQR:* On the Acrobot balancing task, Figure 5, in the presence of Gaussian noise with stddev $= 0.0001$, ILQR is unable to converge with standard finite difference approximations (cost $> 2000$). Algorithms based on $(1,1)$-balanced spinners, $\mathbf{H}$ and $\mathbf{HD}_1$ as well as the product of two independent random $(1,1)$-balanced spinners: $\mathbf{HD}_1\mathbf{HD}_2$, converge within 30 iterations to the solution with cost $< 100$. As we see, introducing randomized versions and adding $\mathbf{HD}$ blocks leads to faster convergence rate. On Cartpole, Figure 6, as for the acrobot task, we observed that $\mathbf{H}$-based approaches are superior to others. In this setting we did not plot the curve for the Hadamard deterministic matrix $\mathbf{H}$ since it is identical as the one for $\mathbf{M} = \mathbf{HD}$ ("Hadamard random" on the plot).

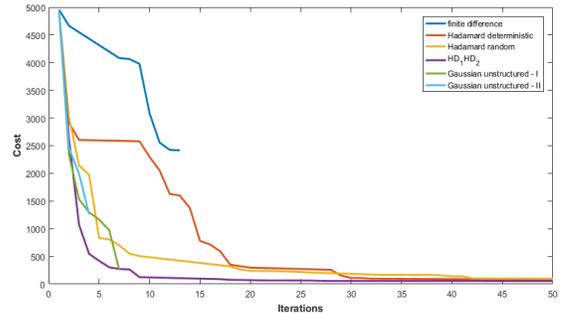

Fig. 5. Comparison of finite difference methods on Acrobot balancing.

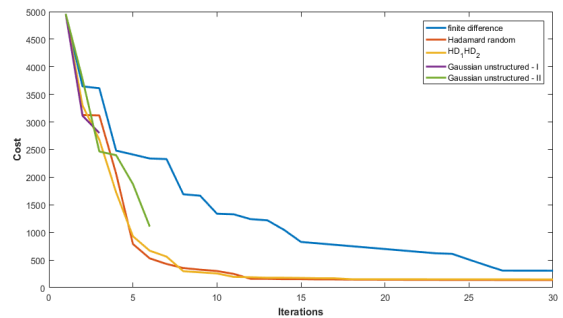

Fig. 6. As in Figure 5, but for the Cartpole task.

## B. Arm Control in Mujoco

In Figure 7, we compare the precision with which the end-effector of a 5-DOF arm simulated in Mujoco can be steered to a goal position using torque-control optimized using TROSS [21], an ADMM based implementation of DDP. For 10 different runs, fixed target, we compared reaching precision obtained with structured finite difference based on random Hadamard matrices **HD** and standard finite differences. The cost is computed as the squared distance from the actual robot's fingertip Cartesian position to the target fingertip Cartesian position. Runs were ordered[1] by the increasing cost plotted in the Figure below. All the runs were conducted in the presence of the additional Gaussian noise with stddev = 0.012. As we see in Figure 7, Hadamard runs are characterized by smaller average error and smaller variance. In the absence of noise both methods were practically indistinguishable (consistently giving cost = 0.0002).

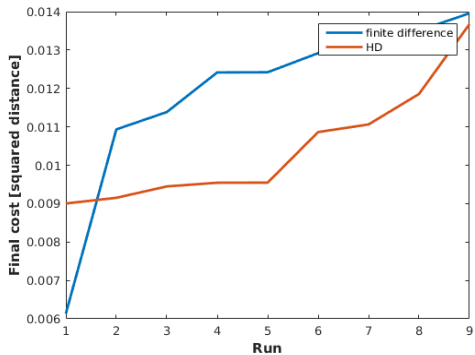

Fig. 7. Arm Control in MujoCo

## IV. QUADRUPED LOCOMOTION

We consider the task of open-loop policy optimization to induce agile walking and turning behaviors for quadruped locomotion on the Minitaur [2] platform, a small-sized, lightweight and dynamic legged quadruped. After system identification, we created a model of the Minitaur in the Bullet simulator [9].

### A. Policy for quadruped locomotion

For these tasks we use a parameterized controller that provides an elliptic motion for the legs. The movement is composed of two sin waves: one for swinging and another one for the extension (length) of the leg. Swing and extension coordinates can be thought as angle and distance in polar coordinate system. For actuation, these coordinates are converted to motor positions for two motors per leg and used with position controllers.

The complete policy is defined by the list of seven parameters. First four parameters define the periodic motion:

- Amplitude vertical ($A_v$) : Amplitude of the extension signal.
- Amplitude horizontal ($A_s$) : Amplitude of the swing signal.
- Speed ($v$) : Frequency of the motion.

[1]we ommitted one optimization failure case

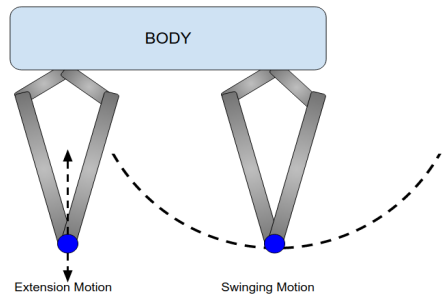

Fig. 8. Leg motion space defined for the locomotion policy.

- Phase vertical ($\phi_v$) : The phase difference between two signals.

Swinging coordinate at specific time $t$ is defined as, $S = A_s \sin(tv)$. Extension coordinate at specific time $t$ is defined as, $V = A_v \sin(tv + \phi)$.

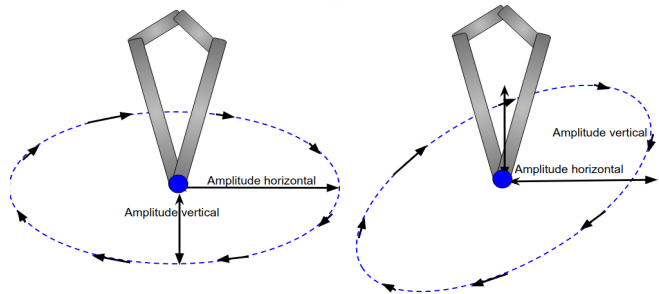

Fig. 9. Example elliptic leg behaviors obtained using the parameters. The first behavior can be obtained when $\phi_v = 0$, the second behavior happens when $\frac{-\pi}{2} < \phi_v < 0$

While all the legs use the same motion (or parameters), each leg has its own phase difference. We use 3 more parameters to represent the phase differences for other 3 legs relative to the first leg. Overall, these seven parameters generate 8 signals (2 per leg) which are then converted to 8 motor angles used by position controllers for the motors.

### B. Task Definition

The problem is defined as episodic: the robot starts each episode from a fixed location. The parameterized policy is applied for some fixed number of steps. The episodes terminate early if the robot reaches the goal destination, or fails to stay relatively parallel to the ground (i.e. fallen to its side).

*Objective and Constraints*: The objective function $f$ is the negated sum of immediate rewards that are hand-designed for inducing running and turning behaviors. Additionally, we impose bound constraints on the parameters: $0 \leq A_v, A_s \leq 0.9, 0.04 \leq v \leq 0.08$ and all phase angles are constrained to be between 0 and $2\pi$.

*Running rewards:* For the running task the immediate reward is defined by the formula:

$$\text{rew} = \alpha d_{\text{forward}} - \beta E + \gamma \delta + \xi h, \tag{9}$$

where $\alpha, \beta, \gamma, \xi$ are fixed positive coefficients and $d_{\text{forward}}, E, \delta, h$ stand for: traversed (signed) distance along the $x$-axis (we force the minitaur to move forward), consumed energy, drift-reward (negated absolute value of the motion along the $y$-axis) and shake-reward (negated absolute value of the movement along the $z$-axis) respectively.

*Turning rewards:* Here, the immediate reward function is even simpler and is of the form

$$\text{rew} = \rho r - \beta E + \xi h, \quad (10)$$

where $r$ stands for the $z$-angular velocity reward (measured as a negated absolute value of the difference between the actual angular velocity and the one we want the robot to achieve) and $\rho$ is another positive coefficient.

### C. Quasi-Newton Optimization and Implicit Filtering

For both turning and walking, bound-constrained policy optimization was done using the popular limited memory quasi-Newton method L-BFGS-B [6], [24]. This approach may be seen as a structured extension of the implicit filtering method [14] which additionally varies the step size $\delta$ in the outer loop to ensure convergence. We seeded Bullet [9] with 100 different starting points (7-dimensional configurations) chosen randomly from the bounded set defined by the constraints put on all the seven parameters. Gradients used by the L-BFGS-B were approximated using deterministic Hadamard matrices.

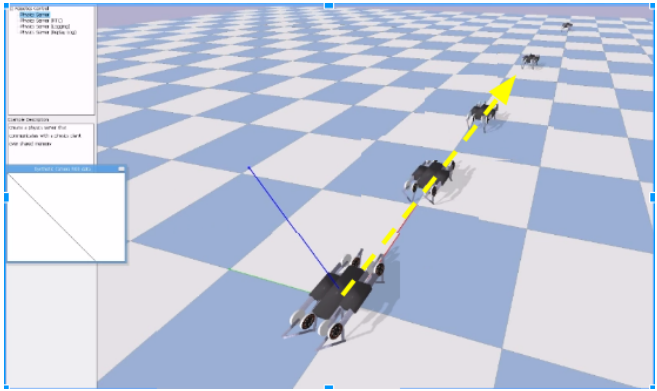

Fig. 10. Learned locomotion behavior snapshots taken from the simulator.

### D. Results

*1) Forward Walking:* For forward locomotion, since the minitaur was constrained to operate within radius rad $= 15.0$, the maximum total reward that can be reached is strictly smaller than 15.0. Using L-BFGS-B, we are able to learn a policy that returns a near-optimal total reward of **14.43** (see Figure 10 for snapshots from Bullet). L-BFGS-B reward improvement as a function of iterations is shown in Figure 11 both for noise free simulations, as well as for multiple runs where actuator noise was injected into the simulated Minitaur. The noise robustness of the optimization can be validated from this experiment.

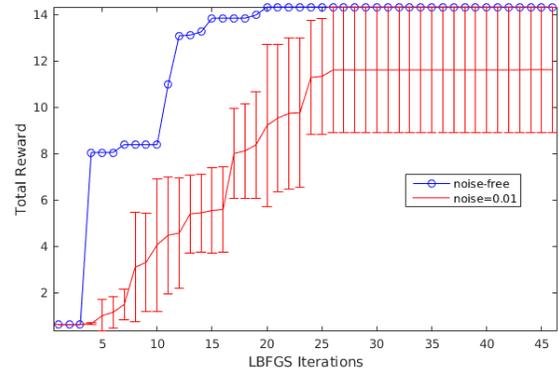

Fig. 11. L-BFGS-B optimization without and with actuator noise.

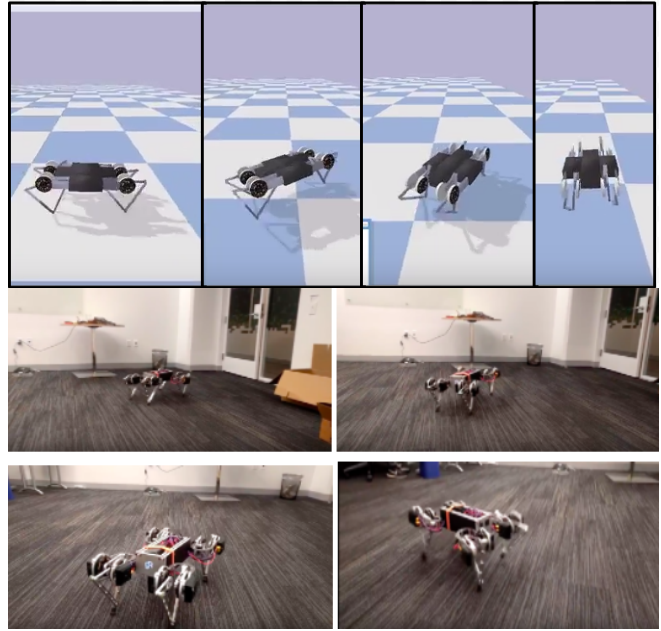

Fig. 12. Learned turning behavior for the minitaur: simulation (top), real (bottom)

*2) Turning Maneuver:* For the turning task, good-quality sets of parameters (with total reward $\text{ret}_{\text{tot}} > -800$) were found within first 100 trials. Interestingly, in that scenario two rotation behaviors were learned: rotating in place and rotating by going along a circle. We also show in Figure 12 that one can compose walking and turning to induce richer navigation behaviors.

*3) Transfer to real hardware:* We tested the learned parameters by deploying them to the hardware platform. The transfer to hardware is relatively successful providing locomotion without any major flaws. On the other hand, we observed that there is a gap between simulation and hardware affecting the speed and direction of the locomotion. This difference is more visible when the robot is tested on different surfaces with different frictional properties such as carpets. Since the policy was not trained for diverse environments, this behavioral difference can be expected. In future work, we are planning to train feedback policies on various environments to obtain more robust behavior. We refer the reader to the accompanying video showing locomotion on the real minitaur.

## V. CONCLUSION

We have demonstrated that structured parameter exploration consistently improves finite difference methods in a number of different settings. A natural extension of the current work is to scale up our methods by exploiting parallelism along the lines of [20] in conjunction with block coordinate updates. We expect this to allow us to train rich, high-dimensional feedback policies in more complex environments. One advantage of finite difference methods is that they can be easily plugged into mature optimization routines, including solvers for general purpose nonlinear programming that can handle complex constraints, that are difficult to explicitly incorporate in existing model-free RL approaches. We intend to explore these themes in future work.

## REFERENCES

[1] http://www.mujoco.org.
[2] https://www.ghostrobotics.io/minitaur.
[3] N. Alon and J. Spencer. *The Probabilistic Method*. John Wiley, 1992.
[4] S. Amaran, N. V. Sahinidis, B. Sharda, and S. J. Bury. Simulation optimization: A review of algorithms and applications. *4OR*, 12(4):301–333, 2014.
[5] M. Bojarski, A. Choromanska, K. Choromanski, F. Fagan, C. Gouy-Pailler, A. Morvan, N. Sakr, T. Sarlós, and J. Atif. Structured adaptive and random spinners for fast machine learning computations. In *Proceedings of the 20th International Conference on Artificial Intelligence and Statistics, AISTATS 2017, 20-22 April 2017, Fort Lauderdale, FL, USA*, pages 1020–1029, 2017.
[6] R. H. Byrd, P. Lu, J. Nocedal, and C. Zhu. A limited memory algorithm for bound constrained optimization. *SIAM Journal on Scientific Computing*, 16(5):1190–1208, 1995.
[7] K. Choromanski, M. Rowland, and A. Weller. The unreasonable effectiveness of structured random orthogonal embeddings. In *to appear in NIPS'17*, 2017.
[8] A. R. Conn, K. Scheinberg, and L. N. Vicente. *Introduction to derivative-free optimization*. SIAM, 2009.
[9] E. Coumans and Y. Bai. Pybullet, a python module for physics simulation for games, robotics and machine learning. http://pybullet.org, 2016–2018.
[10] M. P. Deisenroth, G. Neumann, J. Peters, et al. A survey on policy search for robotics. *Foundations and Trends® in Robotics*, 2(1–2):1–142, 2013.
[11] T. Geng, B. Porr, and F. Wörgötter. Fast biped walking with a reflexive controller and real-time policy searching. In *Advances in Neural Information Processing Systems*, pages 427–434, 2006.
[12] G. H. Golub and C. F. Van Loan. *Matrix computations*, volume 3. JHU Press, 2012.
[13] D. H. Jacobson and D. Q. Mayne. Differential dynamic programming. 1970.
[14] C. T. Kelley. *Implicit filtering*. SIAM, 2011.
[15] N. Kohl and P. Stone. Policy gradient reinforcement learning for fast quadrupedal locomotion. In *Robotics and Automation, 2004. Proceedings. ICRA'04. 2004 IEEE International Conference on*, volume 3, pages 2619–2624. IEEE.
[16] W. Li and E. Todorov. Iterative linear quadratic regulator design for nonlinear biological movement systems.
[17] Y. Nesterov and V. Spokoiny. Random gradient-free minimization of convex functions. *Foundations of Computational Mathematics*, 17(2):527–566, 2017.
[18] A. Y. Ng, A. Coates, M. Diel, V. Ganapathi, J. Schulte, B. Tse, E. Berger, and E. Liang. Autonomous inverted helicopter flight via reinforcement learning. In *Experimental Robotics IX*, pages 363–372. Springer, 2006.
[19] J. Peters and S. Schaal. Policy gradient methods for robotics. In *Intelligent Robots and Systems, 2006 IEEE/RSJ International Conference on*, pages 2219–2225. IEEE, 2006.
[20] T. Salimans, J. Ho, X. Chen, and I. Sutskever. Evolution strategies as a scalable alternative to reinforcement learning. *arXiv preprint arXiv:1703.03864*, 2017.
[21] V. Sindhwani, R. Roelofs, and M. Kalakrishnan. Sequential operator splitting for constrained nonlinear optimal control. In *American Control Conference, 2016*.
[22] Y. Tassa, N. Mansard, and E. Todorov. Control-limited differential dynamic programming. In *Robotics and Automation (ICRA), 2014 IEEE International Conference on*, pages 1168–1175. IEEE, 2014.
[23] S. J. Wright and J. Nocedal. Numerical optimization. *Springer Science*, 35(67-68):7, 1999.
[24] C. Zhu, R. H. Byrd, P. Lu, and J. Nocedal. Algorithm 778: L-bfgs-b: Fortran subroutines for large-scale bound-constrained optimization. *ACM Transactions on Mathematical Software (TOMS)*, 23(4):550–560, 1997.

## VI. APPENDIX

### A. Proof of Theorem 2.1

*Proof:* From the assumptions of the theorem, we get:

$$\mathbf{z} = \mathbf{M}^{-1}\mathbf{m} = \mathbf{M}^{-1}(\mathbf{M}\nabla f(\mathbf{x}_0) + \eta) = \nabla f(\mathbf{x}_0) + \mathbf{M}^{-1}\eta. \quad (11)$$

Therefore it suffices to find an upper bound on $\|\mathbf{M}^{-1}\|_2$. Note that from the assumptions of the theorem:

$$\mathbf{M}\mathbf{M}^\top = \mathbf{D} + \mathbf{N}, \quad (12)$$

where $\mathbf{D}$ is a diagonal matrix with nonzero entries lower-bounded by $\alpha^2 n$ and $\max_{i,j}|N_{i,j}| \leq (1-\beta)\sqrt{n}$. Thus we get:

$$\mathbf{M}(\mathbf{M}^\top \mathbf{D}^{-1} - \mathbf{M}^{-1}) = \mathbf{N}\mathbf{D}^{-1} \quad (13)$$

If we denote by $\sigma_{\min}(\mathbf{M})$ the smallest singular value of $\mathbf{M}$, we then get:

$$\|\mathbf{M}^\top \mathbf{D}^{-1} - \mathbf{M}^{-1}\|_2 \leq \|\mathbf{N}\mathbf{D}^{-1}\|_2 \sigma_{\min}^{-1}(\mathbf{M}) \quad (14)$$

Therefore, by triangle inequality and the fact that the $\|\|_2$ norm is sub-multiplicative, we get

$$\|\mathbf{M}^{-1}\|_2 \leq \frac{\|\mathbf{N}\|_2}{\alpha^2 n \sigma_{\min}(\mathbf{M})} + \frac{\|\mathbf{M}^\top\|_2}{\alpha^2 n} \leq \frac{(1-\beta)\sqrt{n}}{\alpha^2 \sigma_{\min}(\mathbf{M})} + \frac{\|\mathbf{M}^\top\|_2}{\alpha^2 n}, \quad (15)$$

where the last inequality comes from the fact that for $\mathbf{A} \in \mathbb{R}^{n \times n}$: $\|\mathbf{A}\|_2 \leq n \max_{i,j} |A_{i,j}|$. That completes the proof. ∎

### B. Proof of Theorem 2.2

*Proof:* We need to find an upper bound on $\|\mathbf{H}^{-1}\eta\|_\infty$. Since $\mathbf{H}\mathbf{H}^\top = n\mathbf{I}$ and $\mathbf{H}$ is symmetric, we get: $\mathbf{H}\mathbf{D}^{-1} = \frac{1}{n}\mathbf{H}$. Let us focus on a single element of $\frac{1}{n}\mathbf{H}\eta$. Note that it is distributed as: $X = \frac{1}{n}(d_1\eta_1 + ... + d_n\eta_n)$, where $d_i$s are chosen independently and uniformly at random from $\{-1, +1\}$. By Azuma's inequality we get: $\mathbb{P}[|d_1\eta_1 + ... + d_n\eta_n| > t] \leq 2e^{-\frac{t^2}{2\Delta^2}}$, where $\Delta = \|\eta\|_2$. Therefore, from the union bound we get: $\mathbb{P}[\|\mathbf{H}^{-1}\eta\|_\infty \leq \frac{t}{n}] \geq 1 - 2ne^{-\frac{t^2}{2\Delta^2}}$. Now it suffices to take: $t = \Delta\sqrt{g(n)\log(n)}$, notice that $\|\eta\|_2 \leq \sqrt{n}\|\eta\|_\infty$ and that completes the proof. ∎